\documentclass[conference, a4paper]{IEEEtran}
\IEEEoverridecommandlockouts
\usepackage{cite}
\usepackage{url}
\usepackage{amsmath,amssymb,amsfonts}
\usepackage{algorithmic}
\usepackage{graphicx}
\usepackage{textcomp}
\usepackage{xcolor}
\usepackage{multirow}
\usepackage{tablefootnote}
\def\BibTeX{{\rm B\kern-.05em{\sc i\kern-.025em b}\kern-.08em
    T\kern-.1667em\lower.7ex\hbox{E}\kern-.125emX}}
\begin{document}

\title{Prediction of Lane Number Using Results From Lane Detection}

\author{\IEEEauthorblockN{Panumate Chetprayoon}
\IEEEauthorblockA{Mobility Technologies Co., Ltd.\\
Tokyo, Japan}
\and
\IEEEauthorblockN{Fumihiko Takahashi}
\IEEEauthorblockA{Mobility Technologies Co., Ltd. \\
Tokyo, Japan}
\and
\IEEEauthorblockN{Yusuke Uchida}
\IEEEauthorblockA{Mobility Technologies Co., Ltd. \\
Tokyo, Japan}}

\maketitle

\begin{abstract}
The lane number that the vehicle is traveling in is a key factor in intelligent vehicle fields. Many lane detection algorithms were proposed and if we can perfectly detect the lanes, we can directly calculate the lane number from the lane detection results. However, in fact, lane detection algorithms sometimes underperform. Therefore, we propose a new approach for predicting the lane number, where we combine the drive recorder image with the lane detection results to predict the lane number. Experiments on our own dataset confirmed that our approach delivered outstanding results without significantly increasing computational cost.

\end{abstract}

\section{Introduction}
The need for intelligent vehicle system is supported by the fact that humans are the main cause of car accidents \cite{lum1995interactive}. Lane number, which means the number of lane that the vehicle currently belongs to, is an essential factor for intelligent vehicle fields in many aspects such as in road recognition \cite{aufrere2001model}, ego-state tracking \cite{kasprzak1998adaptive}, and ego-lane analysis \cite{berriel2017ego}. More importantly, \cite{popescu2015lane} also emphasized that it is very useful for intelligent vehicle to know the lane number that the vehicle is travelling in. This is because lane number can provide many opportunities such as navigation assistance, environment understanding, and collision assistance systems. As such, developing an accurate lane number prediction module is needed.

Many state-of-the-art lane detection algorithms were proposed such as \cite{neven2018towards} \cite{pan2018SCNN}. If lane detection can work flawlessly, we can get the lane number from the lane detection results without any problems. However, in fact, in some conditions such as low light conditions, lane detection algorithms perform less well than expected \cite{liu2020lane}. Therefore, we came up with the idea that we could utilize the result from lane detection, which is sometimes imperfect, to predict the lane number.

\section{Model}
\label{sec2}
We proposed 4 models with different inputs ($X$). The output ($Y$) is a lane number.

\subsection{Image}
\label{meth_img}
First of all, we handled lane number prediction as a traditional image classification problem so we used image as an input. This model will be used as a baseline to measure how well other models can perform.

\subsection{Lane Binary Mask}
\label{meth_binary}
Next, we considered to use the lane binary mask which is defined as the lane detection result. If we can perfectly detect the lanes, we can directly calculate the lane number from the lane binary mask. To calculate the lane number from the lane binary mask, we count the number of the detected lane lines which has positive slope value.

\subsection{Image + Lane Binary Mask}
\label{meth_img_plus_binary}
\begin{figure}[t]
\centering
\includegraphics[scale=0.28]{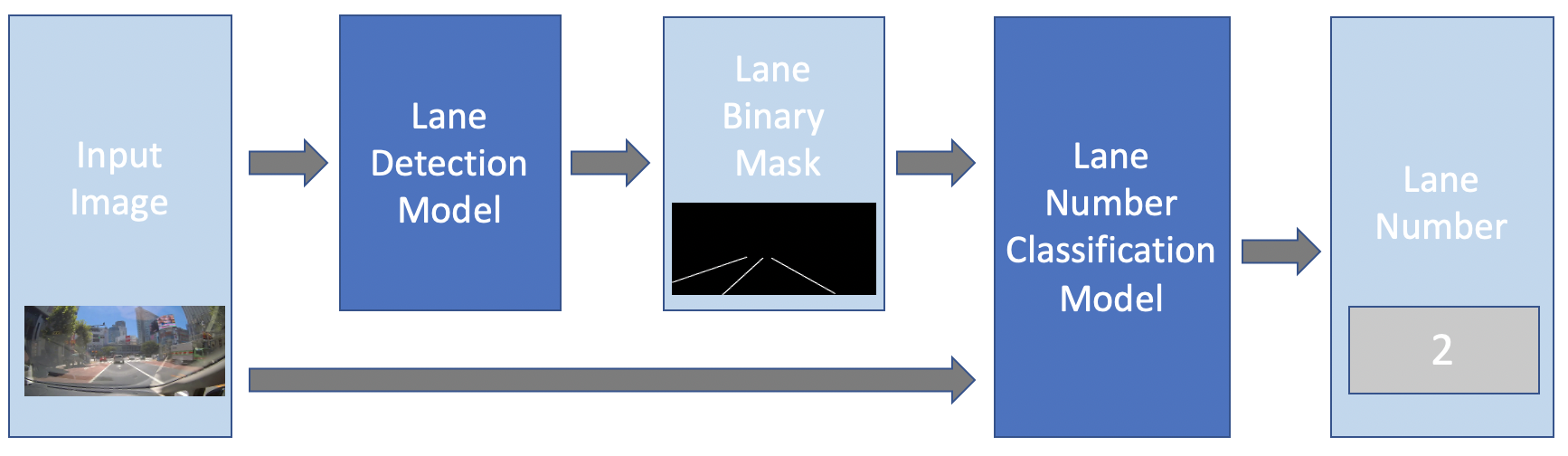}
\caption{Lane number prediction pipeline}
\label{fig_model}
\end{figure}

Lane detection is a method for extracting the information from drive recorder image. However, some information may be lost during the lane detection process. Therefore, our approach prevents this problem by using both drive recorder image and lane binary mask as inputs. The pipeline is shown in Figure~\ref{fig_model}.

\subsection{Image + ($t \pm n$ Lane Binary Mask) }
\label{meth_tpmn}
Since our input is drive recorder video data, it means that we also have other frames data. For example, consider the image at frame $t$ which means the current image, we also have image at frame $t-1$ and $t+1$ which mean the previous frame and the next frame. The main problem of the lane binary mask is that the lane detection module cannot work perfectly due to some conditions. To alleviate this problem, we can utilize the image at frame $t-1$ and $t+1$ as inputs to the lane detection module. We expected that even though the lane detection module failed to predict the lane number at frame $t$, it might still perform well for the image at frame $t-1$ or $t+1$.

\section{Implementation}
\label{sec3}
For dataset, in this experiment, we use our own dataset. Actually, there are some publicly available lane detection datasets. For example, CULane \cite{pan2018SCNN} and TuSimple lane detection \cite{tusimple} datasets are widely used in many academic papers. We can convert lane detection dataset to lane number dataset as we mentioned in Section~\ref{meth_binary}. However, in our study, we did not use these open source lane detection datasets. This is because our approach utilizes the image at frame $t \pm n$ but these datasets do not provide this data.

For lane detection model, we used the excellent off-the-shelf lane detection algorithm named LaneNet \cite{neven2018towards}, and its implementation on GitHub can be found on \cite{maybeshewill}. We first retrain LaneNet by using pretrained model from TuSimple lane detection \cite{tusimple} provided on \cite{maybeshewill}. This lane detection model has 99 million FLOPs and 35 million number of parameters. Our lane detection data consist of 3500 images with 80:20 training and testing.

For lane number classification model used in Section~\ref{meth_img}, Section~\ref{meth_img_plus_binary}, Section~\ref{meth_tpmn}, we have Conv-block defined by Conv($3\times3$) - ReLU - BatchNorm - MaxPooling($2\times2$) - Dropout. Our model starts with Conv 32 ($3\times3$) then Conv-block with size 64,256,64 respectively. Then, flatten with dense-ReLU-Dropout and end with softmax. This lane number classification model has 22 million FLOPs and 3 million number of parameters. 

We resized the input image to $100 \times 100$. Our lane number data consist of 7000 images with 80:20 training and testing. We labeled the data as a classification dataset. $X$ is an input as explained in the Section~\ref{sec2}. $Y$ is the output lane number consists of ${0,1,2,3,4}$. $0$ means the vehicle is in the situation which is hard to determine which lane it currently belongs to such as turning or crossing the crossroads. $4$ means the vehicle belongs to the lane number 4 or more. This is because our data mainly consist of lane number 1, 2, and 3.

\section{Experimental Results}
\label{sec4}

\begin{table}[ht]
\centering
\begin{tabular}{|l|c|c|c|}
\hline
Model             &FLOPs    &Params    &Accuracy \\ \hline
Image             &22M         &3M      &0.78      \\ \hline
Lane Binary Mask   &99M         &35M      &0.80      \\ \hline
Image + Lane Binary Mask   &121M     &38M   &0.87      \\ \hline
Image +  &\multirow{2}{*}{121M}  &\multirow{2}{*}{38M} &\multirow{ 2}{*}{0.90}      \\ 
($t \pm n$ Lane Binary Mask) & & & \\ \hline
\end{tabular}
\caption{Lane Prediction Accuracy}
\label{results}
\end{table}
The result of each model shown in Table~\ref{results} confirmed that our proposed model, described in Section~\ref{meth_img_plus_binary} and Section~\ref{meth_tpmn} (in this case, we used $n=1$), can outperform the baseline described in Section~\ref{meth_img} and Section~\ref{meth_binary}.

In terms of speed and memory, as mentioned in Section~\ref{sec3}, lane number classification model has significantly lower number of parameters and FLOPs compared to lane detection model. So, our proposed model can lead to better result without significantly reducing the performance.

\begin{figure}[t]
\centering
\includegraphics[scale=0.3]{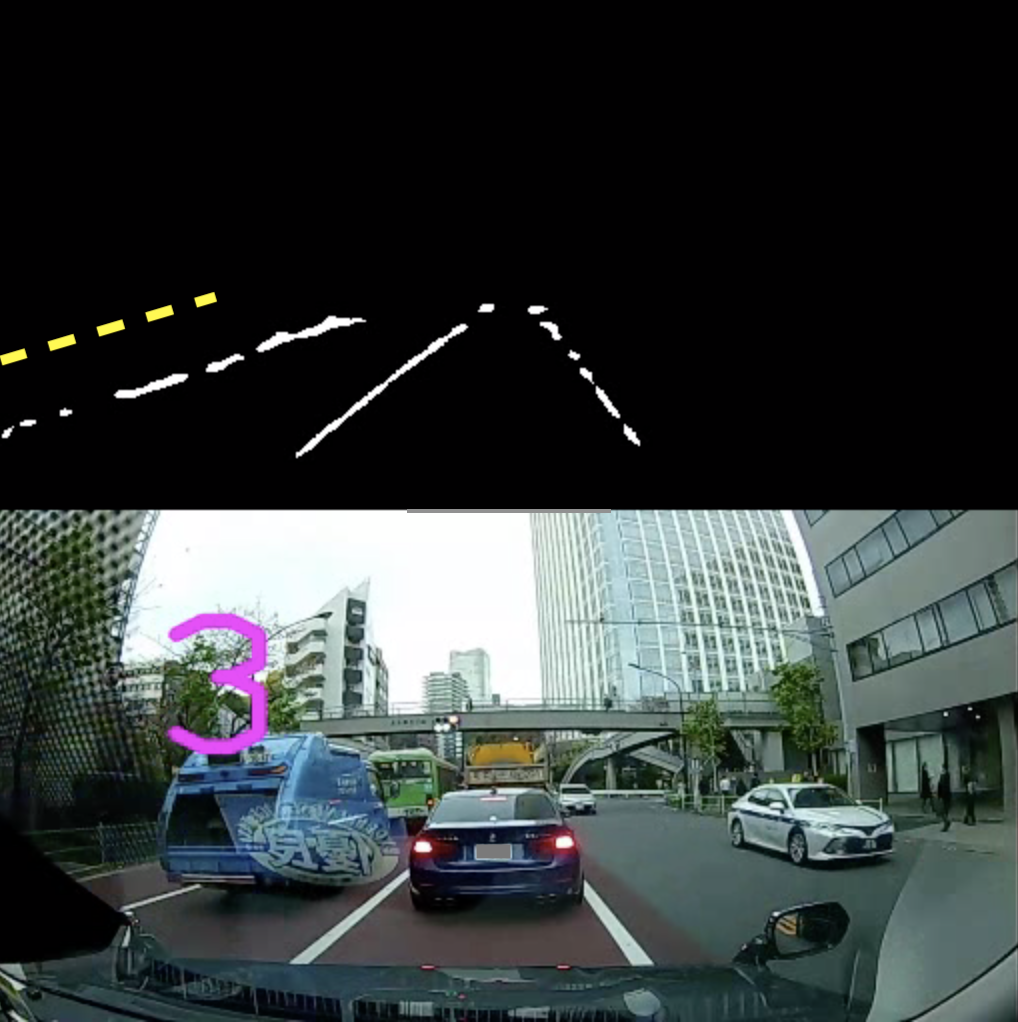}
\caption{Lane number prediction using our approach explained in Section~\ref{meth_img_plus_binary}}
\label{fig_img_and_mask}
\end{figure}

Figure~\ref{fig_img_and_mask} shows the lane detection results (top) and the lane number prediction result (bottom) denoted by purple number shown on the top left corner of the screen. This shows that even though the lane detection algorithm fails to detect the lane binary mask, as we can see that the most left lane line (denoted by the yellow dotted line) is actually not detected, but the final output of lane number prediction is still correct. This is because the model uses input data not only from lane binary mask but also the image. This example, and also the result mentioned afterwards, explicitly confirmed that our approach works very well.

\section{Conclusion}
\label{sec5}
This paper is an attempt to predict the lane number that the vehicle is currently traveling in from drive recorder data. The main contribution of this paper is that our proposed model, combining result from lane detection with drive recorder image, can lead to better result of lane number prediction without significantly increasing computational cost. Experiments on our drive recorder video dataset verified the effectiveness of the proposed model with 90 percent accuracy.

\bibliographystyle{IEEEtran}
\bibliography{sample}

\end{document}